# Predictive Model for Gross Community Production Rate of Coral Reefs using Ensemble Learning Methodologies

Umanandini S, Rishivardhan M, Aouthithiye Barathwaj SR Y, Jasline Augusta J, Shrirang Sapate, Reenasree S, Vigneash M

*Abstract*— **Coral reefs play a vital role in maintaining the ecological balance of the marine ecosystem. Various marine organisms depend on coral reefs for their existence and their natural processes. Coral reefs provide the necessary habitat for reproduction and growth for various exotic species of the marine ecosystem. In this article, we discuss the most important parameters which influence the lifecycle of coral and coral reefs such as ocean acidification, deoxygenation and other physical parameters such as flow rate and surface area. Ocean acidification depends on the amount of dissolved Carbon dioxide ($CO_2$). This is due to the release of $H^+$ ions upon the reaction of the dissolved $CO_2$ gases with the calcium carbonate compounds in the ocean. Deoxygenation is another problem that leads to hypoxia which is characterized by a lesser amount of dissolved oxygen in water than the required amount for the existence of marine organisms. In this article, we highlight the importance of physical parameters such as flow rate which influence gas exchange, heat dissipation, bleaching sensitivity, nutrient supply, feeding, waste and sediment removal, growth and reproduction. In this paper, we also bring out these important parameters and propose an ensemble machine learning-based model for analyzing these parameters and provide better rates that can help us to understand and suitably improve the ocean composition which in turn can eminently improve the sustainability of the marine ecosystem, mainly the coral reefs.**

## I. Introduction

Coral reefs are marine organisms, ecosystems or colonies of Calcareous Organisms, that secrete Calcium Carbonate. Almost all Corals are colonial organisms, composed of hundreds and thousands of individual animals known as Polyps. The coral ecosystem consists of Polyps, Calcareous algae, Shell forming creatures and lime secreting plants. Corals are of two types, soft corals and hard corals, where hard corals have a main contribution to coral reefs.

Coral reefs are calcareous organisms so they need carbon for calcification. They get carbon from a photosynthetic organism called Zooxanthellae. Zooxanthellae require a protected environment and carbon dioxide to perform photosynthesis. Thus, coral reefs form a symbiotic relation with Zooxanthellae, where coral reefs provide a protected environment for Zooxanthellae to perform photosynthesis and Zooxanthellae in turn provides carbon to coral reefs for calcification [1]. The optimal temperature for coral reefs to grow is 23-27 °C but in extreme cases, it can be found at 20-23 °C. So ideally coral reefs are found in the warm water of tropical regions. In tropical regions, areas with warm currents are suitable for coral formation as this region has an ideal temperature. Since these are not found in cold current regions thus, they cannot be found in upwelling zones [2]. So mostly these regions are found on the eastern part of continents except some oceanic currents like Labrador and Oyashio ocean currents, which are cold currents found on eastern regions of North American and Russian continent. Coral reefs are formed in shallow water, ideally, most coral reefs exist in water less than 180 feet. This is because as we go deeper Zooxanthellae is not able to perform photosynthesis due to lack of sunlight. Another condition for coral reefs to exist is that water should be saltish and free from sediments, so ideally coral reefs are not found in freshwater lakes, rivers, silty coasts, muddy mouths, rivers streams, or delta regions. There are three types of coral reefs, Fringing Coral Reefs, Barrier Coral Reefs and Atoll Coral Reefs [3]. According to Darwin's theory of Subsidence, coral reefs are formed in a sequence from Fringing, Barrier to Atoll Coral reefs. The Mesoamerican Reef (MAR) is the largest barrier reef in the western hemisphere, stretching for 625 miles along the coast of Honduras, Guatemala, Belize and Mexico. Florida Coral reef is the third-largest coral reef in the world and the only living coral reef in the United States continent.

Coral Reefs act as a natural barrier against erosion and storm surge. It is the largest biogenic Calcium Carbonate, producer. It provides a substrate for mangroves and acts as a large habitat for plants and animals. Coral reefs play a major role in the food chain of the marine ecosystem. When the water temperature is inappropriate, the coral expels the algae (Zooxanthellae) present in the coral, this turns the coral white. This process is called Coral bleaching. In this case, corals are not dead but they are under more stress. In such cases, Zooxanthellae drops to 60% and photosynthetic pigments also drops to 50-80%.

Anthropogenic activities like chemical pollution, industrial pollution, mechanical damage, nutrient loading, nutrient's sediments offloading and algal bloom, ocean acidification, excessive fertilization are the human threats to the bleaching of coral reefs [4].

Artificial Intelligence is a rapidly growing technology that has created a huge impact in many areas of research. Machine learning is a subdomain of artificial intelligence that involves machines trying to learn to make decisions by learning from data. This learning is done with the help of various algorithms and evaluations to attain maximum accuracy. These algorithms try to learn the pattern from data, mathematically. Using this trained model, evaluations are done to find the accuracy of the predictions. A new metric, weights, is introduced which is adjusted based upon the accuracy of the model. The model training is repeated and the weights are adjusted accordingly, until an optimum accuracy level is obtained.

Machine learning finds a wide range of applications in many areas that includes medical analysis, defense, vehicular technologies, etc. Thus, having valuable data can have a significant impact on the performance of these algorithms. Various insights can be derived when using data that is clean and the appropriate algorithm is used. Ocean data is one such case where the data is vast and it can be utilized to predict the ocean elements, monitor and protect the ocean cover using the machine learning algorithms [5]. In the paper [6], a machine learning framework is implemented to predict the ocean wave height with up to 90% accuracy. These works can significantly help to avoid ocean disasters.

In this paper, a data consisting of parameters such as alkalinity, nutrient level, temperature, pH level are used to predict the Gross Production Rate (GPR) of that marine ecosystem. An analysis is also done to find out which algorithm can work the best in the given dataset and using various metrics such as R2 score, Mean Square Error (MSE), Mean Average Error (MAE), etc.

## II. DEPENDABLE PARAMETERS

### A. Influence of $CO_2$ in pollution and disruption of coral reefs

The $CO_2$ emission has been on an exponential increase in the past two centuries due to various factors such as industrialization, globalization, deforestation and fossil fuel combustion. As per reports, there has been a rise of 36% in $CO_2$ level in the past two decades alone. And one-third of the total $CO_2$ is being absorbed by the oceans. This results in ocean acidification, biodiversity loss, degradation of coral reefs, and limited settlement by coral larvae, impaired coral reproduction and effects such as coral bleaching [7]. These effects occur when the $CO_2$ dissolves in the ocean to form a weak carbonic acid whose bonds easily break resulting in the formation of carbonates and bicarbonates which react with the coral reefs and result in the above-mentioned effects.

The $CO_2$ dissolving in oceans results in acidification of the ocean, this results in a reduction in the saturation amount of crystallized form of $CaCO_3$ i.e., aragonite [8-10]. This acidic condition interferes with the calcification of coral reefs. The carbonic acid formed from the dissolved $CO_2$ is and dissipation of $H^+$ ions occur. Then these dissipated $H^+$ ions react with the carbonate ions resulting in the formation of bicarbonates, which results in the reduction of carbonates ions necessary for calcification. This basically results in the reduction of the growth rate of the coral reefs than it is being destroyed. The way it happens is by disrupting the larval settlement which helps in acquiring the zooxanthellae which are very favorable for coral growth. Thus, declining the chances of coral recovery.

Zooxanthellae play a vital role in gross production of coral i.e., coral reef formation. They bind within the cell walls of the corals and pass on the required nutrients to the corals with help of photosynthesis [11-13]. These zooxanthellae are not only be affected by acidification, but also by increased temperatures. The increased temperatures damage the photosystems of these zooxanthellae which results in the release of Hydrogen peroxides which results in bleaching of these corals as the coral expel these zooxanthellae.

These effects turn this calcification-based corals environment into algal environments [14-15]. When this happens, the coral loses its sustainability in the environment. Also, coral reefs act as a shelter for fishes which preserves the dynamic nature of the coral reefs. If the corals are damaged by these effects, then there is a loss in the biodiversity of these dependent species in and around coral reefs.

### B. Influence of $O_2$ (Ocean Deoxygenation and its Effects on corals) in pollution and disruption of coral reefs.

Ocean deoxygenation is the loss of dissolved oxygen in the marine ecosystem. There has been a steep loss of $O_2$ dissolved in the oceans over the half-century. The main reason for this ocean deoxygenation is due to global climate changes due to global warming and the nutrient enrichment along the coastal lines. This causes pressure in the marine biological environment due to the increase in the biological $O_2$ demand [16-17]. The current-day corals are hard corals that exist based on a symbiotic relationship between the coral and the photosynthetic organism and other supporting organisms with respect to coral tissue and its structural skeleton.

Deoxygenation poses a greater risk to marine organisms especially coral reefs. Hypoxia is referred to as a state when the $O_2$ level required in the tissue is inadequate for the regular physiological functions of the organism. This reduction in $O_2$ at the hypoxia state may be observed through the changes in the biological and physiological response of the affected organisms. This leads to reduced growth and prolonged presence at this state may result in mortality.

The symbiotic relationship holds greater importance in nutrient transfers and reef framework formation acceleration by aiding in calcium carbonate deposition and calcification. This symbiotic relationship contributes to the productivity and tolerance of towards fluctuating temperature, pH and especially deoxygenation.

Hypoxia may occur due to one or more of the following reasons, a higher temperature may decrease the dissolved $O_2$ in water also in the marine ecosystem it also increases the metabolism of the marine organisms consuming an increased amount of oxygen through respiration [18]. Eutrophication requires a greater oxygen demand by microorganisms (BOD). Surprisingly the mortality of marine organisms increases oxygen consumption. As the decomposition of these dead organisms consumes more oxygen for bacterial breakdown. This results in a cascading effect reducing the dissolved oxygen further.

*C. Physical Parameters*

The flow rate affects various processes affecting the growth and life cycle of corals such as gas exchange, heat dissipation, bleaching sensitivity, nutrient supply, feeding, waste and sediment removal, growth and reproduction. The importance of water flow is that it facilitates enhanced gas exchange in corals. The coral's anatomy is very simple hence they are completely dependent on the surrounding environment for gas exchange i.e., diffusion. During the day with help of zooxanthellae's photosynthetic ability the coral releases oxygen and during the night, $CO_2$ has released by respiration [19]. To increase diffusion ability the corals, increase the surface area with the help of pinnules from polyp tentacles. When oxygen accumulates in the Diffusive Boundary Layer (DBL) during the day, the efflux rate of oxygen is reduced as the concentration gradient between the coral tissue and seawater decreases, i.e., the oxygen concentration is high both inside and outside of the coral. Conversely, at night, a depletion of oxygen in the DBL will slow down the influx of oxygen into coral tissue, as the concentration gradient is also reduced, i.e., the oxygen concentration is low both inside and outside of the coral. Thus, the DBL can be regarded as a physical barrier that impedes diffusion. Here when the water flow increases the DBL's thickness around corals decreases. The higher flow rates assist in enhanced diffusion. Flow rate also affects heat dissipation and sediment removal. This in turn affects the overall lifecycle of corals and gross coral production rate i.e., the formation of coral reefs.

## III. MACHINE LEARNING

*A. Multiple Linear Regression*

The linear regression model is used to find the ideal linearity relationship between the different data points in a dataset. This model deals with finding a proportionality between the parameters so as to establish a linear straight graph between them [20]. When the number of features that are considered to build a linear regression, model is more than one, the correlation between the individual features are compared to find the best fitting model. The mathematical model of linear regression is established in eq. (1).

$$y = b + w_1 x_1 + w_2 x_2 ... w_n x_n \quad (1)$$

The model then finds the ideal slope line by minimizing the sum of the squared residuals. From eq. (1), it is clear that the regression coefficients (B1, B2, ..., Bn) of all individual features with respect to the dependent variable are equated to find the final value of the prediction.

*B. Support Vector Regression*

Support Vector Regression (SVR) is a specific type of Support Vector Machine (SVM) wherein a decision boundary is computed to divide the data points into specific groups [21]. A standard line termed as "hyperplane" is then used to predict the regression values. The SVR model attempts to find the best fitting model for a given data by minimizing the distance between the hyperplane and boundary plane. The hyperplane for a given data is chosen by using a kernel function as given in eq. (3) and the dimension of the data is also tuned to find the best fitting model.

$$-a < y - wx + b < +a \quad (2)$$

Various kernels such as Gaussian, RBF, sigmoid, linear and polynomial kernels are usually used in SVR algorithms. Thus, the solution that satisfies eq. (3) is selected as the hyperplane.

*C. Decision Trees*

The Decision Tree Regression (DTR) is a commonly used algorithm that is used more practically in the field of supervised learning [22]. A decision tree is a tree-structured regressor and a classifier which branches the data into nodes and sub nodes based on true/false questions until it reaches a state where it is confident to give out predictions. The branches are divided into nodes by considering factors such as entropy, information gain, etc.

$$E(S) = \sum_{i=1}^{c} -p_i \log_2 p_i \quad (3)$$

The entropy defined in eq. (4) gives the amount of randomness in the data and when the entropy value is more than zero it indicates that the data requires further splitting into nodes. The information gain gives a measure of how easily the data splits into their classes or values. Thus, an ideal attribute in a data is the one with minimal entropy and maximal information gain, which can lead to forming a decision tree model which carries maximum insights from the data.

*D. Random Forests*

The Random Forest Regression is an advancement done in the decision trees algorithm. Overfitting is a prevailing problem in decision trees when the model learns to notice only a few features of the data [23]. This problem is overcome by the use of Random forest algorithm which gives predictions based on multiple decision trees. The Random forest algorithm shows more robustness and rapidity of predictions.

*E. Ridge Regression*

Ridge Regression is a modified form of linear regression in which the fitting slope line is calculated as given in eq. (5).

$$L = \sum_{i=1}^{n} (y_i - Xw)^2 + \lambda \sum_{j=1}^{p} w_j^2 \quad (5)$$

In Linear Regression, the slope with minimal sum of squared residuals is selected as the best fitting slope for the data whereas in ridge regression, an additional term which is the product of the square of slope and a term "lambda" is added to the cost function to provide the model's fitting line [24]. The term lambda is usually fine-tuned in order to achieve the least variance.

### F. Lasso Regression

LASSO is an acronym that stands for Least Absolute Shrinkage and Selection Operator. It is a statistical algorithm for regularizing data models and selecting features. When a regression model employs the L1 Regularization technique, it is referred to as Lasso Regression. L1 regularization adds a penalty equal to the absolute value of the coefficient's magnitude. This type of regularization can lead to sparse models with few coefficients. Some coefficients may be zeroed out and thus removed from the model. Larger penalties result in coefficient values that are closer to zero.

$$COST(W(t)) = \sum_{i=1}^{N}\left\{y_i - \sum_{j=0}^{M} w_j x_{ij}\right\}^2 + \lambda \sum_{j=0}^{M}|w_j|$$

### G. Elastic Regression

Elastic net is a popular type of regularized linear regression that combines two well-known penalties, the L1 and L2 penalty functions. Elastic Net is a linear regression extension that includes regularization penalties in the loss function during training. One common penalty is to penalize a model using the sum of the squared coefficient values. This is known as an L2 penalty. An L2 penalty reduces the size of all coefficients while preventing them from being removed from the model. Another common penalty is to punish a model based on the sum of its absolute coefficient values. This is known as the L1 penalty. An L1 penalty reduces the size of all coefficients while also allowing some coefficients to be reduced to zero, removing the predictor from the model. A penalized linear regression model called an elastic net incorporates both L1 and L2 penalties during training.

$$ElasticNet = \sum_{i=1}^{n}(y_i - y(x_i))^2 + \alpha \sum_{j=1}^{p}|w_j| + \alpha \sum_{j=1}^{p}(w_i)^2$$

### H. Ensemble Learning

Ensemble learning is a supervised learning method that utilizes the predictions from more than one model to average out the predictions. This method is more preferred since it does not depend on a single algorithm which is essential in cases where the insights from a particular model does not take into account all the valuable insights in a data [25]. The different methods used in ensemble learning includes bagging, boosting, stacking, etc. In the bagging method, the data is split and the portioned data is fed into different individual learning algorithms and the mean of their predictions gives the final value. Boosting method involves allocating weights when an individual model gives a right prediction so that it tries to give more positive predictions. The former method helps to reduce the variance error while the latter aims to reduce the bias error. We adopt the bagging technique for ensemble learning of the proposed system. Bagging is basically a bootstrap aggregation technique that follows a parallel training ideology. In this type, the same data is trained in different models and then aggregated as a single model. In this type, we can reduce the variance of the entire data and also can be implemented very easily.

$$f_{bag} = f_1(X) + f_2(X) + ... + f_b(X) \tag{6}$$

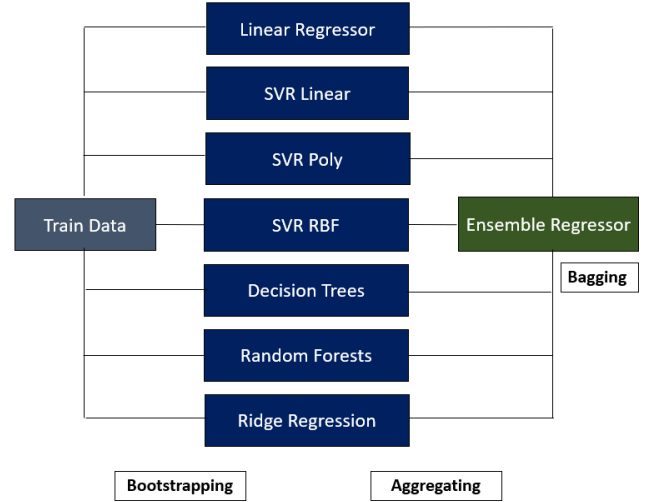

Figure 1: Block diagram of ensemble learning

## IV. EXPERIMENT AND RESULTS

### A. Dataset

The examination of the dependency of nutrient pollution and pH on coral reefs has been carried out for years. In this dataset, the dependency of elevated nitrate ($NO_3$) and phosphate ($PO_4^3$) on net community calcification (NCC) and net community production (NCP) rates of individual taxa and combined reef communities are acquired. Based on the observation and calculations, several dependable parameters that include Tank Total Alkalinity, Tank Temperature, Tank pH, Tank Phosphate, Nitrate, Silicate, Tank $CO_2$, Tank $hCO_3$, Tank $CO_3$, Tank Dissolved Inorganic Carbon, Tank Aragonite Saturation State Tank Calcite Saturation State, Tank $pCO_2$, Tank $fCO_2$ Residence time, Flowrate, Surface area Ash Free Dry Weight, Dry Weight Day/Night, Net Community Clarification Rate, Net Community Production Rate, Respiration, Gross Community Production Rate. The tank header observation values are dropped from the dataset as they do not significantly contribute to the production rate. In order to reduce the computation load, calculation-based parameters are dropped from the data set that include $pCO_2$, $fCO_2$ Net Community Clarification Rate, Gross Community Clarification Rate, Dry Weight, etc.

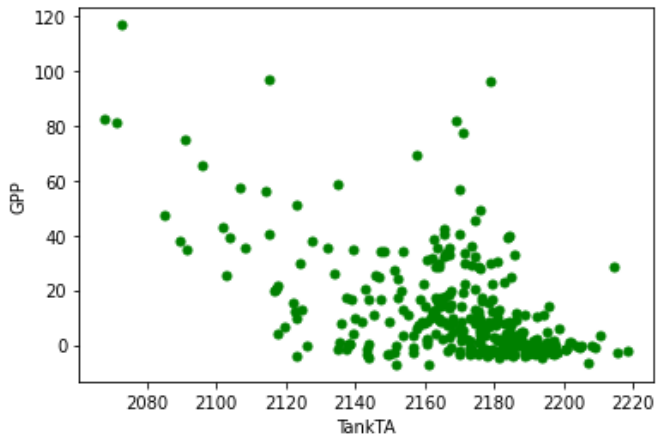

Figure 2: Data distribution of total alkalinity in the tank with respect to Gross Production Rate

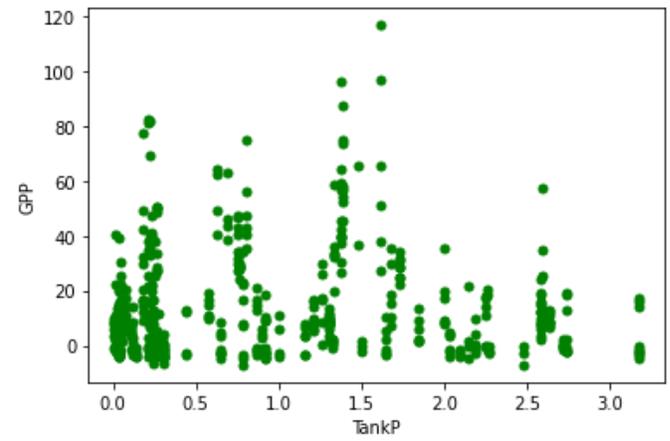

Figure 5: Data distribution of Phosphate in the tank with respect to Gross Production Rate

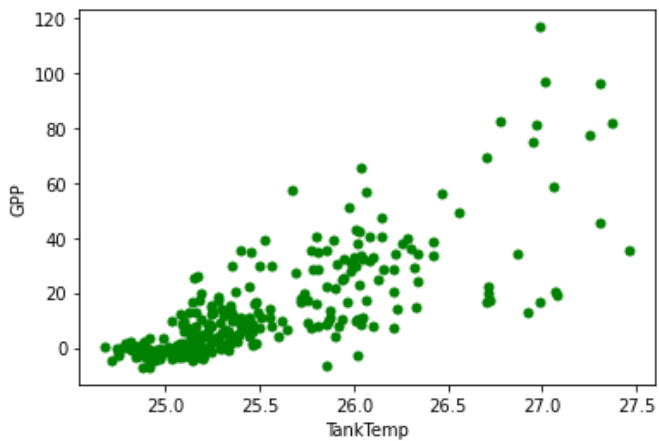

Figure 3: Data distribution of temperature in the tank with respect to Gross Production Rate

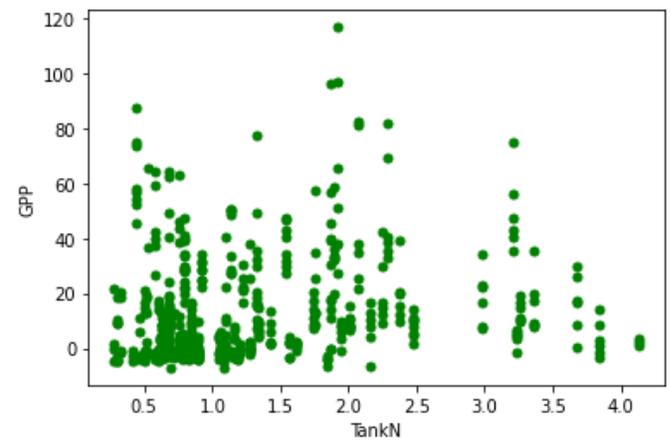

Figure 6: Data distribution of Nitrate in the tank with respect to Gross Production Rate

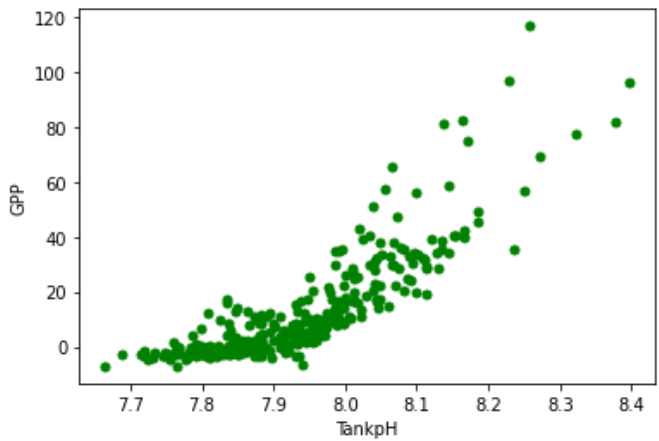

Figure 4: Data distribution of pH in tank with respect to Gross Production Rate

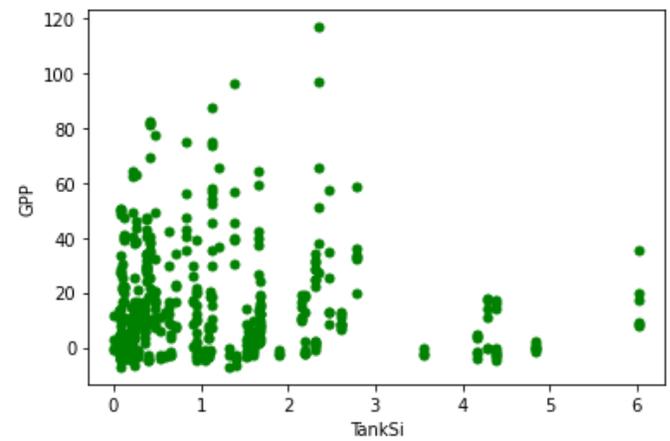

Figure 7: Data distribution of Silicate in the tank with respect to Gross Production Rate

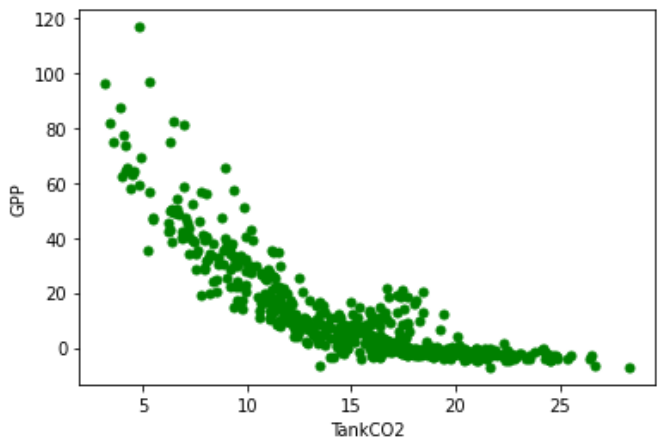

*Figure 8: Data distribution of CO2 in the tank with respect to Gross Production Rate*

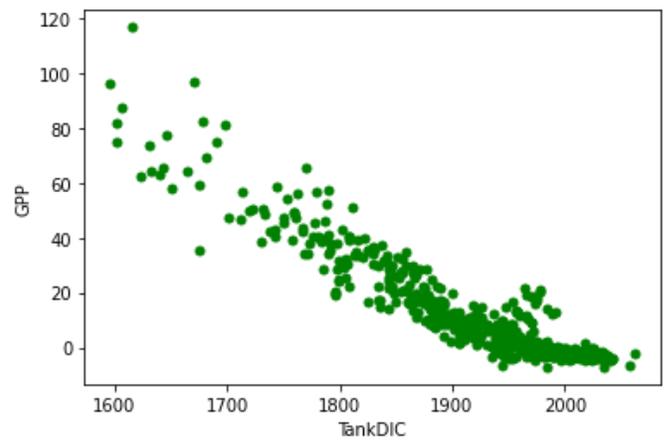

*Figure 11: Data Distribution of Dissolved Inorganic Carbon in the tank with respect to Gross Production Rate*

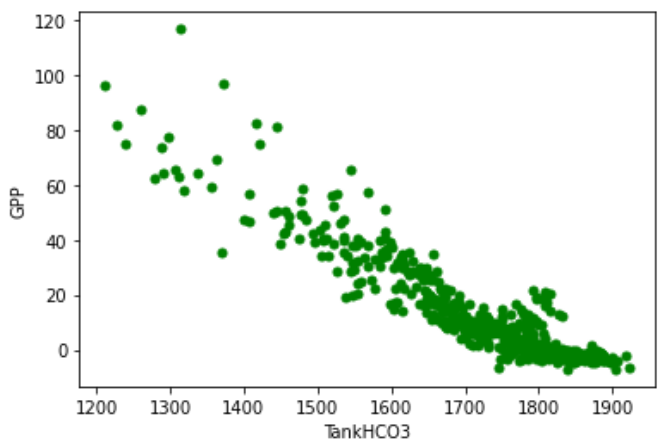

*Figure 9: Data distribution of HCO3 in the tank with respect to Gross Production Rate*

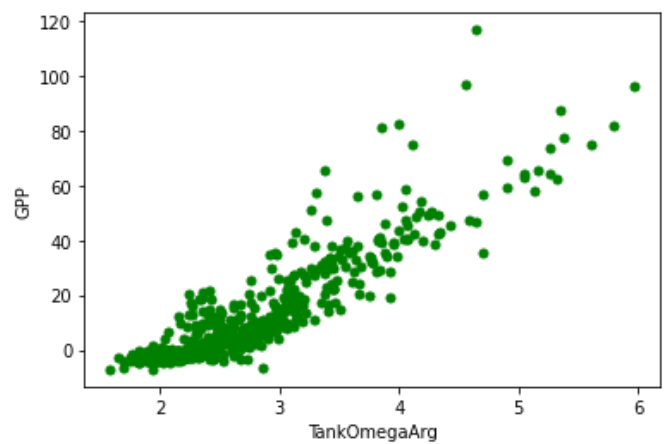

*Figure 12: Data distribution of Aragonite Saturation State in the tank with respect to Gross Production Rate*

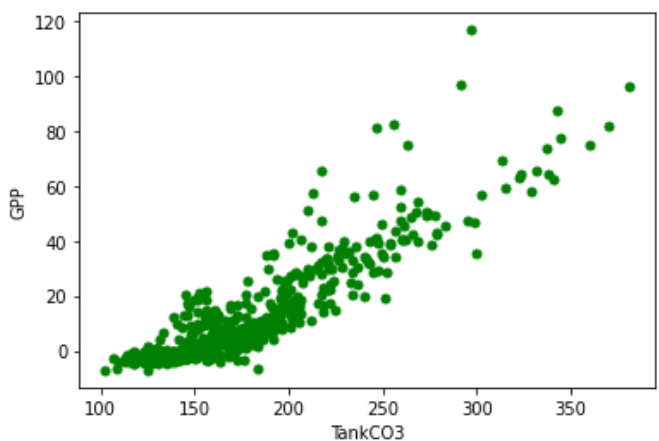

*Figure 10: Data distribution of CO3 in the tank with respect to Gross Production Rate*

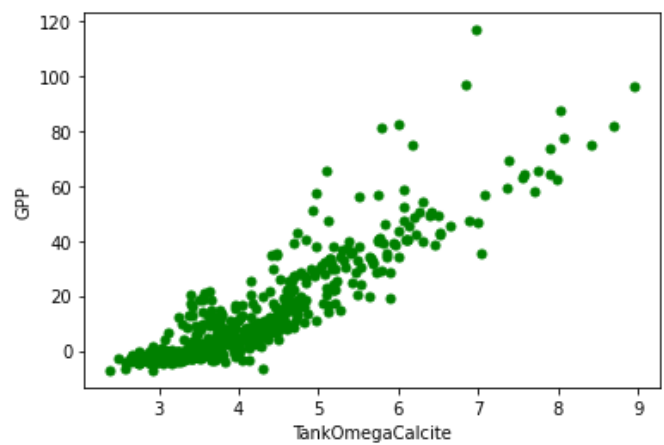

*Figure 13: Data distribution of Calcite Saturation State in the tank with respect to Gross Production Rate*

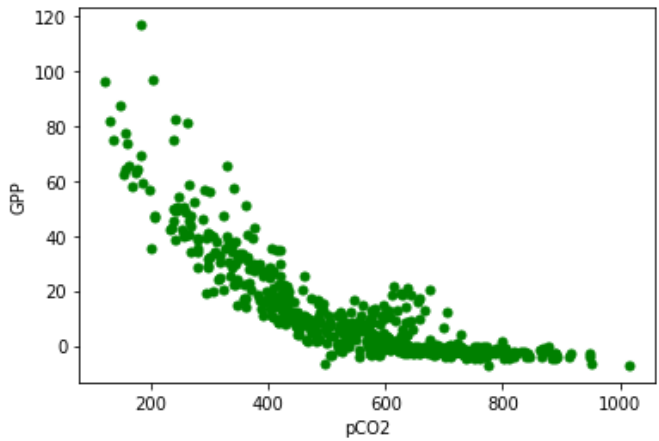

*Figure 14: Data distribution of pCO2 in the tank with respect to Gross Production Rate*

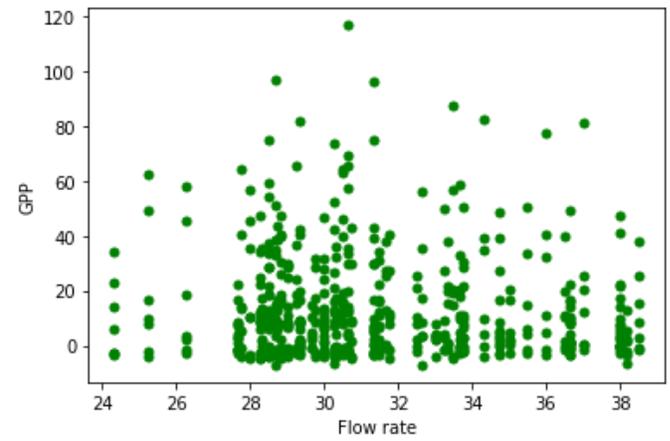

*Figure 17: Data distribution of Mean Flow Rate with respect to Gross Production Rate*

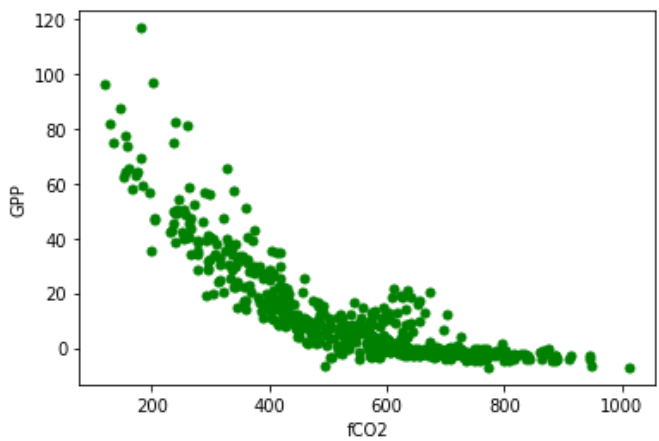

*Figure 15: Data distribution of fCO2 in the tank with respect to Gross Production Rate*

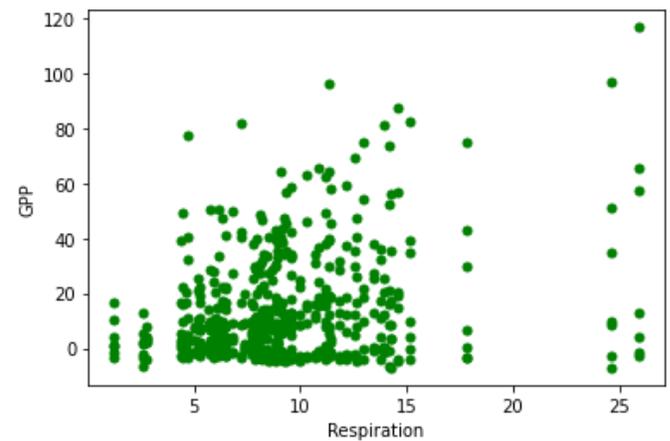

*Figure 18: Data distribution of Respiration with respect to Gross Production Rate*

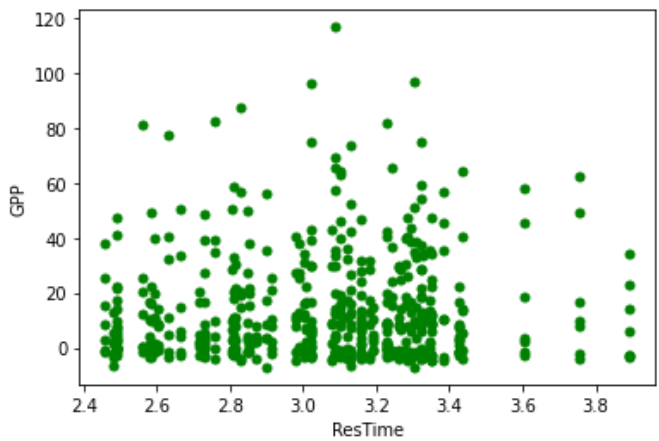

*Figure 16: Data distribution of Residence Time with respect to Gross Production Rate*

### B. Training

Training of the model is executed in the Google Colaboratory environment using Nvidia G4 Tesla 16GB GPU, 8GB memory with Core i5-8300H 2.30GHz Quad core CPU. Out of 505 data instances, 60% of data is contributed for the training process and the remaining 40% is contributed for testing process. Scikit learn package is used for training majority of the models. Additionally, NumPy and pandas are used for numerical computation and data visualization processes.

### C. Results

The evaluation metrics used here include regression score, mean squared error and mean absolute error. Out of the 7 proposed models, i.e., linear regression, support vector regression with linear kernel, support vector regression with polynomial kernel, support vector regression with radial basis function, decision trees, random forests and ridge regression, the best performing algorithm in this dataset was found to be linear regression followed by ridge regression. The bagging ensemble model proves the same. The test regression score is said to be 0.94 for the linear regression model. The complete test evaluation data is given in the table 1.a

TABLE 1: MODELS WITH EVALUATION METRICS

| Algorithm / Metric | R2 | MSE | MAE |
|---|---|---|---|
| Linear Regression | 0.909065 | 34.925414 | 4.453529 |
| SVR Linear | 0.826017 | 66.821746 | 5.359787 |
| SVR Poly | 0.380073 | 238.09649 | 9.582600 |
| SVR RBF | -0.01309 | 389.10001 | 14.85956 |
| Decision Trees | 0.766164 | 89.809518 | 6.249231 |
| Random Forests | 0.885570 | 43.949093 | 4.857569 |
| Ridge Regression | 0.844423 | 59.752677 | 5.469316 |
| Lasso Regression | 0.867730 | 53.178910 | 4.328352 |
| Elastic Regression | 0.831238 | 49.540124 | 4.469316 |

## V. CONCLUSION AND FUTURE WORKS

Coral reefs host the global highest biodiversity in an ecosystem. Though the gross area of the coral reefs covering the masses is very less i.e., <0.1% of the ocean they house various exotic marine organisms. Apart from these reefs also provides necessary food and prevents flooding thus helping in the marine and supporting human life. Surprisingly coral reefs support more than half a billion people either supporting economically or directly for their daily survival. They act as indicators of global climate changes. The proposed ensemble machine learning model follows the bagging technique. This technique is widely used in environmental analysis due to its simplicity and less variance over large data. Out of all the models used in this proposed works, linear regression tends to outperform all the other algorithms in a rationally larger margin. With the help of this model, we can easily analyze the best parameters that are to be worked on to improve the gross production of coral into coral reefs. Thus we conclude with this that, with ensemble model, we can find the most important parameters which could lay a major role in combating climate change.